\documentclass{ifacconf}
\usepackage[T1]{fontenc}
\usepackage[latin9]{inputenc}
\usepackage{amsmath}
\usepackage{subfig}
\usepackage{amssymb}
\usepackage{graphicx}      
\usepackage{natbib}        
\usepackage{balance}

\begin{document}
\begin{frontmatter}

\title{Reachability as a Unifying Framework for Computing Helicopter Safe
Operating Conditions and Autonomous Emergency Landing\thanksref{footnoteinfo}} 

\thanks[footnoteinfo]{This research was supported in part by the Navy Innovative Science and Engineering program under Grant 219WFD-SG-18-003 and Grant N00421-19-WX00648 and in part by the Office of Naval Research under Grant N00014-19-WX00546.}

\author[First,Second]{Matthew R. Kirchner} 
\author[Third]{Eddie Ball} 
\author[Third]{Jacques Hoffler}
\author[Third]{Don Gaublomme}

\address[First]{Image and Signal Processing Branch, Code D5J1000, Naval Air Warfare Center Weapons Division, China Lake, CA, USA (e-mail: matthew.kirchner@navy.mil).}
\address[Second]{Electrical and Computer Engineering Department, University of California, Santa Barbara, CA 93106, USA (e-mail: kirchner@ucsb.edu).}
\address[Third]{Flight Vehicle Modeling and Simulation Branch, Naval Air Warfare Center Aircraft Division, Patuxent River, MD, USA (e-mail: \{eddie.ball, jacques.hoffler, donald.gaublomme\}@navy.mil).}

\begin{abstract}                
We present a numeric method to compute the safe operating flight conditions
for a helicopter such that we can ensure a safe landing in the event
of a partial or total engine failure. The unsafe operating region
is the complement of the backwards reachable tube, which can be found
as the sub-zero level set of the viscosity solution of a Hamilton\textendash Jacobi
(HJ) equation. Traditionally, numerical methods used to solve the
HJ equation rely on a discrete grid of the solution space and exhibit
exponential scaling with dimension, which is problematic for the high-fidelity
dynamics models required for accurate helicopter modeling. We avoid
the use of spatial grids by formulating a trajectory optimization
problem, where the solution at each initial condition can be computed
in a computationally efficient manner. The proposed method is
shown to compute an autonomous landing trajectory from any operating
condition, even in non-cruise flight conditions. 
\end{abstract}


\end{frontmatter}

\section{Introduction}
This paper presents a method to efficiently compute unsafe operating
flight conditions for which no control sequence exists to safely land
the helicopter in the event of partial or total engine failure. A
chart called a height-velocity diagram or H-V diagram needs to be
produced for each airframe, and a human pilot can utilize the diagram
to avoid operating in unsafe conditions. If operating in the safe
region, there exists a control sequence to initiate an autorotation,
whereby the helicopter enters a glide slope so as to maintain rotor
inertia and then flare to slow down near the ground and land safely
\cite[Chapter 11]{FAAHandbook}. H-V diagrams are ultimately determined
through flight testing, which is inherently dangerous since the test
objective is to define the unsafe boundaries of flight operations.
Therefore, there is a need to accurately compute the safe and unsafe
regions directly from the helicopter dynamics, and construct a H-V
diagram prior to flight testing. The capability to produce H-V diagrams
is desired for aircraft that are in the early design stages and have
not yet flown, aircraft that are already flying and preparing for
an H-V flight test event, as well as refining the H-V diagram of operational
aircraft. The capability is required for aircraft that do not have
a representative, high-fidelity flight simulation, as well as for
more mature aircraft for which a high-fidelity simulation is available.

We consider the safe region as any initial flight condition where
there exists a control sequence that can steer the system to a safe
landing condition, i.e. minimal vertical and horizontal velocity at
the ground level, rotor near level, etc. Determining the set of states
of a dynamical system that can be driven into a particular final condition
is commonly referred to as reachability analysis, and reachable sets
can be determined from the sub-zero level set of the viscosity solution
to a Hamilton\textendash Jacobi (HJ) partial differential equation
(PDE) (\cite{mitchell2005time}).

Traditionally, these HJ PDEs are solved numerically by constructing
a dense discrete grid of the solution space (\cite{mitchell2008flexible}),
and are supported by mature theory (\cite{osher2006level}). Despite
this, HJ reachability analysis has suffered one critical draw-back:
Computing the elements of a spatial grid scales poorly with dimension,
and therefore have limited applicability for vehicle problems where
the dimension of the state space is greater than four. The work of
\cite{harno2018safe} attempted to compute helicopter safe operating
regions using the grid-based method of \cite{mitchell2008flexible},
but was restricted to a two-dimensional state space, and therefore
is not representative of actual helicopter motion and is not applicable
for safety critical use.

Recent research based on generalizations of the Hopf formula (\cite{hopf1965generalized})
have avoided a spatial grid of the state space and instead form a
grid over \emph{time}, where numerical solutions are obtained by constructing
a trajectory optimization problem. These techniques were used in \cite{darbon2016algorithms}
to compute solutions to the HJ equation for systems with state and
time-independent Hamiltonians of the form $\dot{x}=f\left(u\left(t\right)\right)$,
and \cite{kirchner2018primal} expanded the classes of systems
to general linear systems in high-dimensions. Additionally, these
methods were successfully applied to vehicle control problems such
as collaborative pursuit-evasion (\cite{kirchner2017time}) and coordination
of heterogeneous groups of vehicles (\cite{kirchner2020}).

The methods described above based on the Hopf formula can be seen
as an optimization problem based on the costate trajectory of the
system. We propose in this work to formulate an optimization of the
state trajectory directly, which allows consideration of the non-linear
dynamics encountered in helicopter motion. There is no need to directly
formulate the optimal Hamiltonian of the system, since the Hamiltonian
frequently resulting from helicopter dynamics does not have a standard
form.

There have been various proposals to compute the H-V diagram using
trajectory optimization (see e.g. \cite{johnson1977helicopter,lee1988optimal,carlson2006h,yomchinda2013real,bibik2012helicopter}),
but they suffer several drawbacks. These methods minimize cost functionals
with weighted cost terms and a terminal state constraint, which do
not guarantee necessary ground landing conditions are met. Therefore,
the solution of these optimization problems do not, in general, give
the reachable set. Since the functional is parameterized by the terminal
state at precisely terminal time, $t$, time is an unknown free parameter
that must be solved for. When combined with the terminal constraint, it
is not guaranteed that a candidate trajectory is feasible for a particular
time and numeric convergence issues can be observed.

We propose to construct an implicit surface representation of the
desired safe landing condition and, in doing so, create a trajectory
optimization that gives the backwards reachable tube (\cite{bansal2017hamilton}).
We consider the case where the time-to-land need not be known a priori
by considering that the intersection of the helicopter with the ground
could occur at any time on the interval $\left[0,\infty\right)$.
We utilize Lagrange polynomials to create a collocation scheme that
converges rapidly with the number of time samples and can represent
the required half-infinite time intervals.
From this we compute the optimal landing sequence to autonomously
achieve a safe landing from any initial condition that is within the
backwards reachable tube.

\section{Reachability and Safe Operating
Regions}

We consider helicopter dynamics
\begin{equation}
\frac{d}{ds}x\left(s\right)=f\left(x\left(s\right),u\left(s\right)\right),\label{eq:general dynamics}
\end{equation}
for $s\in\left[0,t\right]$ where $x\in\mathbb{R}^{n}$ is the system
state and $u\left(s\right)\in\mathcal{U}\subset\mathbb{R}^{m}$ is
the control input, constrained to lie in the admissible control set
$\mathcal{U}$. We denote by $\gamma$ the state trajectory $\left[0,t\right]\ni s\mapsto$
$\gamma\left(s;x,u\left(\cdot\right)\right)\in\mathbb{R}^{n}$ that
evolves in time with measurable control input $\left[0,s\right]\mapsto u\left(\cdot\right)\in\mathcal{U}$
according to $\left(\ref{eq:general dynamics}\right)$ starting from
initial state $x$ at $s=0$. The trajectory $\gamma$ is a solution
of $\left(\ref{eq:general dynamics}\right)$ in that it satisfies
$\left(\ref{eq:general dynamics}\right)$ almost everywhere:
\begin{align}
\frac{d}{ds}\gamma\left(s;x,u\left(\cdot\right)\right) & =f\left(\gamma\left(s;x,u\left(\cdot\right)\right),u\left(s\right)\right),\label{eq:dynamic constraints}\\
\gamma\left(0;x,u\left(\cdot\right)\right) & =x.\nonumber 
\end{align}
We denote by $\Theta_{0}\subseteq\mathbb{R}^{n}$ as the goal set
that represents the set of acceptable safe landing conditions. We
seek to determine if a control sequence exists that drives the system
into $\Theta_{0}$ at exactly time $t$. The set of all initial states
where there exists a control to drive the system to the set at exactly
time $t$ is called the \emph{backward reachable set} (BRS) of the
system and is defined as
\begin{equation}
\Theta_{S}\left(t\right)=\left\{ x:\exists u\left(\cdot\right)\in\mathcal{U},\gamma\left(t;x,u\left(\cdot\right)\right)\in\Theta_{0}\right\} .\label{eq: formal definition of BRS}
\end{equation}

\subsection{The Connection of the BRS to the HJ Equation}

We represent the set of safe landing conditions, $\Theta_{0}$, implicitly
with the function $J:\mathbb{R}^{n}\rightarrow\mathbb{R}$ such that
\begin{equation}
\Theta_{0}=\left\{ x\in\mathbb{R}^{n}|J\left(x\right)\leq0\right\} \label{eq:implicit surface def}
\end{equation}
and use it to construct a cost functional for the system trajectory
$\gamma\left(s;x,u\left(\cdot\right)\right)$, given terminal time
$t$ as 
\begin{equation}
K\left(t,x,u\left(\cdot\right)\right)=\int_{0}^{t}\mathcal{I}_{\mathcal{U}}\left(u\left(s\right)\right)ds+J\left(\gamma\left(t;x,u\left(\cdot\right)\right)\right),\label{eq: Cost Function}
\end{equation}
where the running cost function $\mathcal{I}_{\mathcal{U}}:\mathbb{R}^{m}\rightarrow\mathbb{R}\cup\left\{ +\infty\right\} $
is the characteristic function for the set $\mathcal{U}$ and is defined
by
\[
\mathcal{I}_{\mathcal{U}}\left(u\right)=\begin{cases}
0 & \text{if}\,u\in\mathcal{U}\\
+\infty & \text{otherwise.}
\end{cases}
\]
The value function $\varphi:\mathbb{R}^{n}\times\left[0,t\right]\rightarrow\mathbb{R}$
is defined as the minimum cost, $K$, among all admissible controls
for a given state $x$ as
\begin{equation}
\varphi\left(x,t\right)=\underset{u\left(\cdot\right)\in\mathcal{U}}{\text{inf}}\,K\left(t,x,u\left(\cdot\right)\right).\label{eq: Value function}
\end{equation}
The value function in $\left(\ref{eq: Value function}\right)$ satisfies
the dynamic programming principle (\cite{evans10})
and also satisfies the following initial value Hamilton\textendash Jacobi
(HJ) equation with $\varphi$ being the viscosity solution of
\begin{equation}
\begin{cases}
\frac{\partial\varphi}{\partial s}\left(x,s\right)+H\left(x,\nabla_{x}\varphi\left(x,s\right)\right)=0,\\
\varphi\left(x,0\right)=J\left(x\right),
\end{cases}\label{eq:Initial value HJ PDE}
\end{equation}
for $s\in\left[0,t\right]$, where the Hamiltonian $H:\mathbb{R}^{n}\times\mathbb{R}^{n}\rightarrow\mathbb{R}\cup\left\{ +\infty\right\} $
is defined by
\begin{equation}
H\left(x,p\right)=\underset{u\in\mathcal{U}}{\text{sup}}\left\langle -f\left(x,u\right),p\right\rangle ,\label{eq: Basic Hamiltonian definition}
\end{equation}
where we denote by $p\in\mathbb{R}^{n}$ the costate variable.
\begin{fact}[\cite{Mitchell_toolbox}]
The zero sub level sets of the viscosity solution $\varphi\left(x,t\right)$
is an implicit surface representation of the finite time backwards
reachable set, $\Theta_{S}\left(t\right)$ defined in $\left(\ref{eq: formal definition of BRS}\right)$.
\end{fact}
We note here the important distinction that the BRS in $\left(\ref{eq: formal definition of BRS}\right)$
defines the set of initial states that can be driven into the set
$\Theta_{0}$ at precisely time $t$. It is possible for the system
to be driven into the set $\Theta_{0}$ at an earlier time $\tilde{t}<t$
and then later exit the set $\Theta_{0}$.
This motivated the characterization of the backwards reachable tube
in \cite{mitchell2005time}.

\subsection{The Backwards Reachable Tube}

We are instead interested if the system can be driven into $\Theta_{0}$
at \emph{any} time on the range $[0,t]$. This set of initial states,
coined the \emph{backwards reachable tube }(BRT) in \cite{bansal2017hamilton}, is defined by
\begin{equation}
\Theta_{T}\left(t\right)=\left\{ x:\exists u\left(\cdot\right)\in\mathcal{U},\exists s\in\left[0,t\right],\gamma\left(s;x,u\left(\cdot\right)\right)\in\Theta_{0}\right\} .\label{eq:formal definition BRT}
\end{equation}
The seminal work of \cite{mitchell2005time} showed that the backwards
reachable tube is found from the zero sub level set of $\varphi\left(x,t\right)$,
which is the viscosity solution to the following modified Hamilton\textendash Jacobi
equation
\begin{equation}
\begin{cases}
\frac{\partial\varphi}{\partial s}\left(x,s\right)+\max\left(0,H\left(x,\nabla_{x}\varphi\left(x,s\right)\right)\right)=0,\\
\varphi\left(x,0\right)=J\left(x\right),
\end{cases}\label{eq:mitchell HJ equation fro BRT}
\end{equation}
with $H\left(x,p\right)$ the same as given above in $\left(\ref{eq: Basic Hamiltonian definition}\right)$.
The intuition is that the value function cannot decrease\footnote{In \cite{mitchell2005time}, a minimum operator is used since time
is defined in that work on the interval $\left[-t,0\right]$.}, which does not allow the level sets to contract. It can be seen
that any trajectory that enters the set $\Theta_{0}$ is not allowed
to escape by ``freezing'' it in time. The reachable set and reachable
tube are connected through the following relation \cite[Proposition 1]{mitchell2007comparing}
\begin{equation}
\Theta_{T}\left(t\right)=\bigcup_{s\in\left[0,t\right]}\Theta_{S}\left(s\right).\label{eq:Tube is union of sets}
\end{equation}
 The boolean set operation in $\left(\ref{eq:Tube is union of sets}\right)$
implies a corresponding property (\cite{osher2006level}) of the implicit
surface representations
\begin{equation}
\varphi_{T}\left(x,t\right)=\underset{s\in\left[0,t\right]}{\min}\varphi_{S}\left(x,s\right).\label{eq:implicit surface rep. of union property}
\end{equation}
The backwards reachable tube, $\Theta_{T}\left(t\right)$, characterizes
the \emph{controllably safe} regions of flight. Therefore, the set
of \emph{unsafe} states is found from
\[
\Theta^{C}:=\underset{t\rightarrow\infty}{\lim}\mathbb{R}^{n}\setminus\Theta_{T}\left(t\right).
\]

\section{Computing the Reachable Tube
Through Optimization}

Under a mild set of regularity conditions \cite[Ch. 7, p. 63]{subbotin1995generalized},
$\varphi\left(x,s\right)$ is the unique viscosity solution of $\left(\ref{eq:Initial value HJ PDE}\right)$
\cite[Th. 8.1, p. 70]{subbotin1995generalized}. The uniqueness of
the solution, $\varphi\left(x,s\right)$, implies that the viscosity
solution is equivalent to the value function and can be found by minimizing
$\left(\ref{eq: Cost Function}\right)$ with constraints given by
$\left(\ref{eq:dynamic constraints}\right)$. The HJ of the BRT in
$\left(\ref{eq:mitchell HJ equation fro BRT}\right)$ can be constructed
by augmenting the dynamics of $\left(\ref{eq:general dynamics}\right)$
with
\begin{equation}
\frac{d}{ds}x\left(s\right)=g\left(x\left(s\right),u\left(s\right),a\left(s\right)\right):=a\left(s\right)f\left(x\left(s\right),u\left(s\right)\right),\label{eq:augmented dynamics}
\end{equation}
where $\left[0,t\right]\ni s\mapsto a\left(s\right)\in\left[0,1\right]$ is measurable scalar function (\cite{mitchell2005time}).
We denote by $\zeta$ the state trajectory of the augmented system
$\left(\ref{eq:augmented dynamics}\right)$ that satisfies
\begin{align}
\frac{d}{ds}\zeta\left(s;x,u\left(\cdot\right),a\left(\cdot\right)\right) & =g\left(\zeta\left(s;x,u\left(\cdot\right),a\left(\cdot\right)\right),u\left(s\right),a\left(s\right)\right),\nonumber \\
\zeta\left(0;x,u\left(\cdot\right),a\left(\cdot\right)\right) & =x.\label{eq:augmented dynamic constraints}
\end{align}
Note that since $a$ is constrained to $\left[0,1\right]$, a value
of $a=1$ makes the augmented dynamics of $\left(\ref{eq:augmented dynamics}\right)$
equivalent to the original dynamics of $\left(\ref{eq:general dynamics}\right)$
and if $a=0$ the dynamics stop completely. We denote by
\[
\sigma\left(s\right):=\int_{0}^{s}a\left(\ell\right)d\ell,
\]
the pseudotime variable, and denote by $\sigma^{\dagger}$ as a quasi-inverse
of $\sigma$ in the sense that
\[
\sigma^{\dagger}\left(\sigma\left(s\right)\right)=s.
\]
The formal definition of this quasi-inverse is given in \cite[Lemma 6]{mitchell2005time},
and it was shown that the trajectories of $\left(\ref{eq:augmented dynamic constraints}\right)$
have the following relations
\[
\gamma\left(\sigma\left(s\right);x,u\left(\sigma^{\dagger}\left(\cdot\right)\right)\right)=\zeta\left(s;x,u\left(\cdot\right),a\left(\cdot\right)\right),
\]
for any $s\in\left[0,t\right]$ \cite[Lemma 4]{mitchell2005time},
and consequently $\zeta$ visits only a subset states of the $\gamma$
\cite[Corollary 5]{mitchell2005time}. Therefore, we can solve the
modified optimization problem to find the reachable tube:
\begin{equation}
\begin{cases}
\underset{u\left(\cdot\right),a\left(\cdot\right)}{\min} & J\left(\zeta\left(t;x,u\left(\cdot\right),a\left(\cdot\right)\right)\right)\\
\text{Subject to} & \frac{d}{ds}\zeta\left(s\right)=g\left(\zeta\left(s\right),u\left(s\right),a\left(s\right)\right)\\
 & \zeta\left(0\right)=x\\
 & u\left(\cdot\right)\in\mathcal{U}\\
 & a\left(\cdot\right)\in\left[0,1\right]\\
 & \dot{a}\left(\cdot\right)\leq0,
\end{cases}\label{eq:modified optimization}
\end{equation}
where the last line of $\left(\ref{eq:modified optimization}\right)$
is an optional regularization term since the optimal $a\left(\cdot\right)$
is not unique.

\subsection{Time-Domain Transformation}

We use collocation to approximate the
trajectory $\zeta$ and first perform a time domain transformation
which maps $s\in\left[0,\infty\right)$ into the computation interval
$\tau\in\left[-1,1\right)$ with the invertible transform defined as
\begin{equation}
s=\Gamma^{-1}\left(\tau\right)=c\log\left(\frac{2}{1-\tau}\right),\label{eq:time domain transformation}
\end{equation}
where the choice of $\Gamma$ is a scaled version of that proposed
in \cite{garg2011pseudospectral}. The dynamics $\left(\ref{eq:augmented dynamics}\right)$
are similarly transformed with
\begin{align*}
\frac{d}{d\tau}x\left(\tau\right) & =\frac{d}{d\tau}\Gamma\left(\tau\right)g\left(x\left(\tau\right),u\left(\tau\right),a\left(\tau\right)\right),
\end{align*}
and with the transform defined in $\left(\ref{eq:time domain transformation}\right)$
becomes
\[
\frac{d}{d\tau}x\left(\tau\right)=\frac{c}{\left(1-\tau\right)}g\left(x\left(\tau\right),u\left(\tau\right),a\left(\tau\right)\right).
\]
Hereafter, we assume that the trajectory, $\zeta$, is a solution
to the transformed dynamics satisfying
\begin{align*}
\frac{d}{d\tau}\zeta\left(\tau;x,u\left(\cdot\right),a\left(\cdot\right)\right) & =\frac{c}{\left(1-\tau\right)}\\
 & \times g\left(\zeta\left(\tau;x,u\left(\cdot\right),a\left(\cdot\right)\right),u\left(\tau\right),a\left(\tau\right)\right)\\
\zeta\left(-1;x,u\left(\cdot\right),a\left(\cdot\right)\right) & =x.
\end{align*}

\subsection{Finite Trajectory Approximation}

We propose a polynomial trajectory approximation, which is commonly
referred to as pseudospectral optimal control, and was introduced
in \cite{elnagar1995pseudospectral} and later refined in \cite{ross2012review}
and \cite{garg2010unified}. We denote by $\zeta^{N}$ as the approximation
to $\zeta$ with Lagrange polynomials as
\begin{equation}
\zeta\left(\tau\right)\approx\zeta^{N}\left(\tau\right)=\sum_{j=0}^{N}x_{j}L_{j}\left(\tau\right),\label{eq: traj approx as poly}
\end{equation}
defined by a set of $N$ points, $x_{j}=\zeta\left(\tau_{j};x,u\left(\cdot\right),a\left(\cdot\right)\right)$,
sampled on the time grid 
\[
\pi^{N}=\left\{ \tau_{j}:j=0,\ldots,N\right\} ,
\]
where each $\tau_{j}\in\left[-1,1\right]$. The Lagrange basis functions
are given as
\[
L_{j}\left(\tau\right):=\prod_{\underset{k\neq j}{k=0}}^{N}\frac{\tau-\tau_{k}}{\tau_{j}-\tau_{k}},
\]
and it follows that
\[
\frac{d}{d\tau}\zeta^{N}\left(\tau_{i}\right)=\sum_{j=0}^{N}x_{j}\dot{L}_{j}\left(\tau_{i}\right)=\sum_{j=0}^{N}D_{ij}x_{j},
\]
where we denote by $D\in\mathbb{R}^{N\times\left(N+1\right)}$ as
the Gauss differentiation matrix constructed with each element, with
row $i$ and column $j$, given by 
\begin{equation}
D_{ij}:=\dot{L}_{j}\left(\tau_{i}\right).\label{eq:diff of lagrange}
\end{equation}
We select $\pi^{N}$ as Legendre-Gauss-Radau (LGR) points with $\pi^{N}$
and the corresponding differentiation matrix, $D$, and quadrature
weights, $w$, are found from \cite[Section 3.3]{shen2011spectral}.
Note that since LGR points do not include a point at $\tau=1$, we
avoid a singularity in $\left(\ref{eq:time domain transformation}\right)$
at $t\rightarrow\infty$. We denote by $D_{0}\in\mathbb{R}^{N\times1}$
as the first column of $D$ corresponding to the boundary condition
and denote by $D_{I}\in\mathbb{R}^{N\times N}$ as the remaining columns
representing the interior nodes such that
\begin{equation}
D=\left[\begin{array}{cc}
D_{0} & D_{I}\end{array}\right].\label{eq:diff matrix partition}
\end{equation}

We denote by $X\in\mathbb{R}^{n\cdot N}$ as the concatenated vector
of all collocation points for $j=1,\ldots,N$ given as
\begin{equation}
X:=\left(x_{1},\cdots,x_{N}\right)^{\top},\label{eq: concat state}
\end{equation}
and likewise denote $U\in\mathbb{R}^{m\cdot\left(N+1\right)}$ as
the concatenated vector of all the collocated control input points
\[
U:=\left(u_{0},\cdots,u_{N}\right)^{\top},
\]
where each $u_{j}$ is the control input at each time $\tau_{j}\in\pi^{N}$.
And similarly, for the augmented inputs, $a_{j}$, we have
\[
A:=\left(a_{0},\cdots,a_{N}\right)^{\top}.
\]
Recall that $x_{0}=x=\zeta\left(-1;x,u\left(\cdot\right),a\left(\cdot\right)\right)$.
We denote by $G\left(X,U,A\right)$ as the concatenated vector of
the equations of motion,
\[
G\left(X,U,A\right):=\left[\begin{array}{c}
a_{0}\frac{c}{\left(1-\tau_{0}\right)}f\left(x,u_{0}\right)\\
a_{1}\frac{c}{\left(1-\tau_{1}\right)}f\left(x_{1},u_{1}\right)\\
\vdots\\
a_{N}\frac{c}{\left(1-\tau_{N}\right)}f\left(x_{N},u_{N}\right)
\end{array}\right],
\]
 evaluated at each point in $\tau_{j}\in\pi^{N}$. It follows the
differentiation matrix $\left(\ref{eq:diff matrix partition}\right)$
of the concatenated of state $\left(\ref{eq: concat state}\right)$
is
\[
\mathcal{D}_{I}=D_{I}\otimes I_{n},
\]
where $\otimes$ denotes the Kronecker product and $I_{n}$ is the
$n\times n$ identity matrix. Likewise, we have
\[
\mathcal{D}_{0}=D_{0}\otimes I_{n}.
\]
The dynamic equality constraint is now written as
\[
\mathcal{D}_{s}X+\mathcal{D}_{0}x=G\left(X,U,A\right).
\]
Recall that $w$ is the vector of LGR quadrature weights. It was proposed
in \cite{garg2010unified} to estimate the terminal state with
\[
x_{N+1}=x+\left(w^{\top}\otimes I_{n}\right)G\left(X,U,A\right).
\]

With $x=\zeta\left(0;x,u\left(\cdot\right),a\left(\cdot\right)\right)$
as the given initial condition, we construct the non-linear programming
problem (NLP): 
\begin{equation}
\begin{cases}
\underset{X,U,A}{\min} & J\left(x_{N+1}\right)\\
\text{Subject to} & \mathcal{D}_{s}X+\mathcal{D}_{0}x=G\left(X,U,A\right)\\
 & x_{N+1}=x+\left(w^{\top}\otimes I_{n}\right)G\left(X,U,A\right)\\
 & u_{j}\in\mathcal{U}\\
 & a_{j}\in\left[0,1\right]\\
 & a_{j}-a_{j-1}\leq0,
\end{cases}\label{eq:reach tube nlp}
\end{equation}
which has a numerical solution that approximates $\left(\ref{eq:modified optimization}\right)$
when $N$ is sufficiently large. The time of minimum cost in $\left(\ref{eq:implicit surface rep. of union property}\right)$
can be found from the following quadrature:
\begin{equation}
t^{*}=w^{\top}\left[\begin{array}{c}
a_{0}\frac{c}{\left(1-\tau_{0}\right)}\\
\vdots\\
a_{N}\frac{c}{\left(1-\tau_{N}\right)}
\end{array}\right].\label{eq:opt time}
\end{equation}

\section{Helicopter Dynamics}

We define the state vector as $x=\left(y,v,\Omega,h,z,P\right)^{\top}\in\mathbb{R}^{6}$
with the vertical velocity, $y$; horizontal velocity, $v$; rotor
angular speed, $\Omega$; height above ground level, $h$; horizontal
displacement, $z$; and engine power, $P$. The control inputs are
given as $u$=$\left(C_{T},\theta\right)^{\top}\in\mathbb{R}\times\left[0,2\pi\right)$
with thrust coefficient, $C_{T}$, effectively the collective control,
and $\theta$, the aircraft pitch angle. Following \cite{lee1986optimal},
the dynamics are as follows:
\begin{equation}
f:\begin{cases}
\dot{y}=g-\frac{\rho}{m}\left(\pi R^{2}\right)\left(\Omega R\right)^{2}C_{T}\cos\theta+\frac{1}{2}\frac{\rho}{m}f_{eh}V_{f}y\\
\dot{v}=\frac{\rho}{m}\left(\pi R^{2}\right)\left(\Omega R\right)^{2}C_{T}\sin\theta-\frac{1}{2}\frac{\rho}{m}f_{ez}V_{f}v\\
\dot{\Omega}=\frac{1}{I_{R}\Omega}\left(P-\frac{1}{\eta}\rho\left(\pi R^{2}\right)\left(\Omega R\right)^{3}C_{p}\right)\\
\dot{h}=-y\\
\dot{z}=v\\
\dot{P}=-\frac{1}{\kappa}P
\end{cases}\label{eq: heli eom}
\end{equation}
where $\kappa$ denotes the engine response time constant, $\eta$
the rotor power efficiency factor, $m$ the mass of the helicopter, $V_f$ the velocity magnitude of the fuselage, $\sigma_R$ the rotor solidity, $R$ the rotor radius, $I_R$ the rotor inertia, $\rho$ the density of the air, and $g$
the acceleration due to gravity. $f_{eh}$ and $f_{ez}$ is the flat plat drag area in the vertical and horizontal directions, respectively. $C_{p}$ denotes the power coefficient
defined as
\[
C_{p}=\frac{1}{8}\sigma_R c_{d}+C_{T}\lambda,
\]
where $c_d$ is the drag coefficient of the rotor airfoil and 
\[
\lambda=\frac{v\sin\theta-y\cos\theta+\nu}{\Omega R}
\]
 is the inflow ratio. The advance velocity, $U_{T}$, is given by
\[
U_{T}=v\cos\theta+y\sin\theta.
\]
The variable $\nu$ is induced velocity and we use the inflow model
\[
\nu=\Omega R\sqrt{\frac{1}{2}C_{T}},
\]
which is the ideal inflow in hover. More sophisticated inflow models
exist but are outside the scope of this work, and the reader is encouraged
to read \cite{johnson1977helicopter} and \cite{chen1986influence}
for more details. 

\section{Results}

We ensure safe landing when the conditions $\left|y\right|\leq y_{\max}$
and $\left|v\right|\leq v_{\max}$ are met as the helicopter is sufficiently
close to the ground; in this case $\left|h\right|\leq1$. The parameters
$y_{\max}$ and $v_{\max}$ are found from the structural specifications
of the airframe. We select the function $J$ that satisfies $\left(\ref{eq:implicit surface def}\right)$
the terminal conditions with
\[
J\left(x\right)=\max\left(\left|y\right|-y_{\max},\left|v\right|-v_{\max},\left|h\right|-1\right),
\]
where, for this example, we chose $y_{\max}=8\,ft/s$ and $v_{\max}=6\,knots$.
The remaining initial states are chosen as the trim conditions such
that $\dot{y},\dot{v}=0$, and initial engine power was set $P_{0}=0$,
simulating an instant engine failure and providing a safety factor
as it gives the largest unsafe set. The input $C_{T}$ is bounded
by a blade stall condition with $C_{T}\leq0.15\sigma_R$ (\cite{lee1988optimal}),
where $\left|\theta\right|\leq40\textdegree$.
The rest of the coefficients for the model in $\left(\ref{eq: heli eom}\right)$
are from \cite[Table 2-1, Table A-1]{yomchinda2013real}.
The number of interior time samples, $N$, is fixed at $24$ and the constant
in $\left(\ref{eq:time domain transformation}\right)$ is set to
$c=3$. To take advantage of the inherent sparsity
of the formulated optimization, we use the NLP code IPOPT (\cite{wachter2006implementation}),
where the constraint Jacobian was computed using automatic differentiation
by the methods of \cite{Andersson2019}.

\begin{figure}
\begin{centering}
\includegraphics[width=8cm]{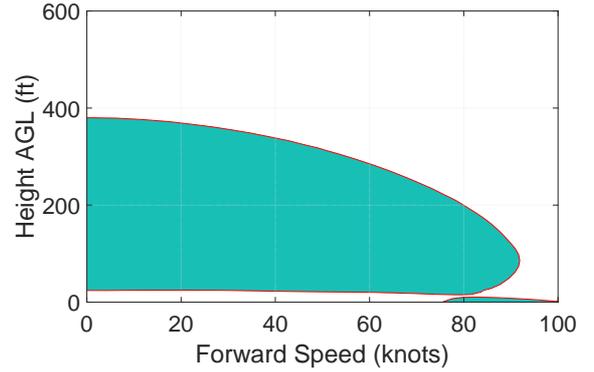}
\par\end{centering}
\caption{\label{fig:H-V}Shown in green is the set of unsafe H-V flight conditions
as computed from the method of Section 3.}
\end{figure}

The value function for the system was computed using the method of
Section 3, and an H-V diagram is
produced by finding the unsafe regions, which is all areas where the
value function is greater than zero. Figure \ref{fig:H-V} shows the
H-V diagram for the dynamics given in $\left(\ref{eq: heli eom}\right)$,
where dark regions on the diagram represent initial height above ground
level (AGL) and forward speed such that a safe landing in the event
of an engine failure is impossible. Note that a disjoint lobe of the
unsafe set appears on the bottom right of Figure \ref{fig:H-V}
and is known as the high-speed unsafe region.

\subsection{Autonomous Autorotation}

\begin{figure*}
\begin{centering}
\subfloat[Height profile during landing.]{\centering{}\includegraphics[width=5.5cm]{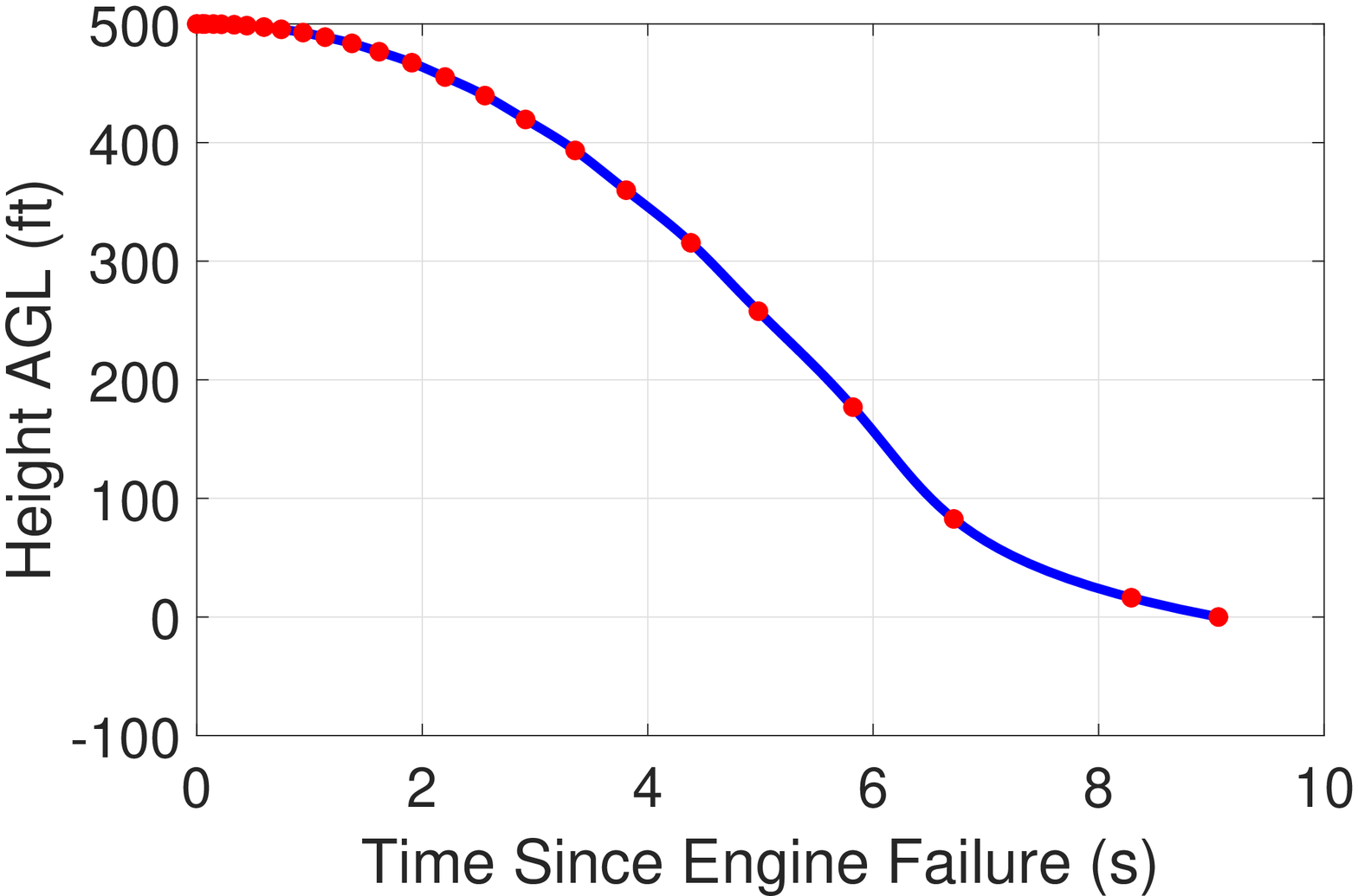}}\hspace*{\fill}\subfloat[Descent rate profile during landing.]{\begin{centering}
\includegraphics[width=5.5cm]{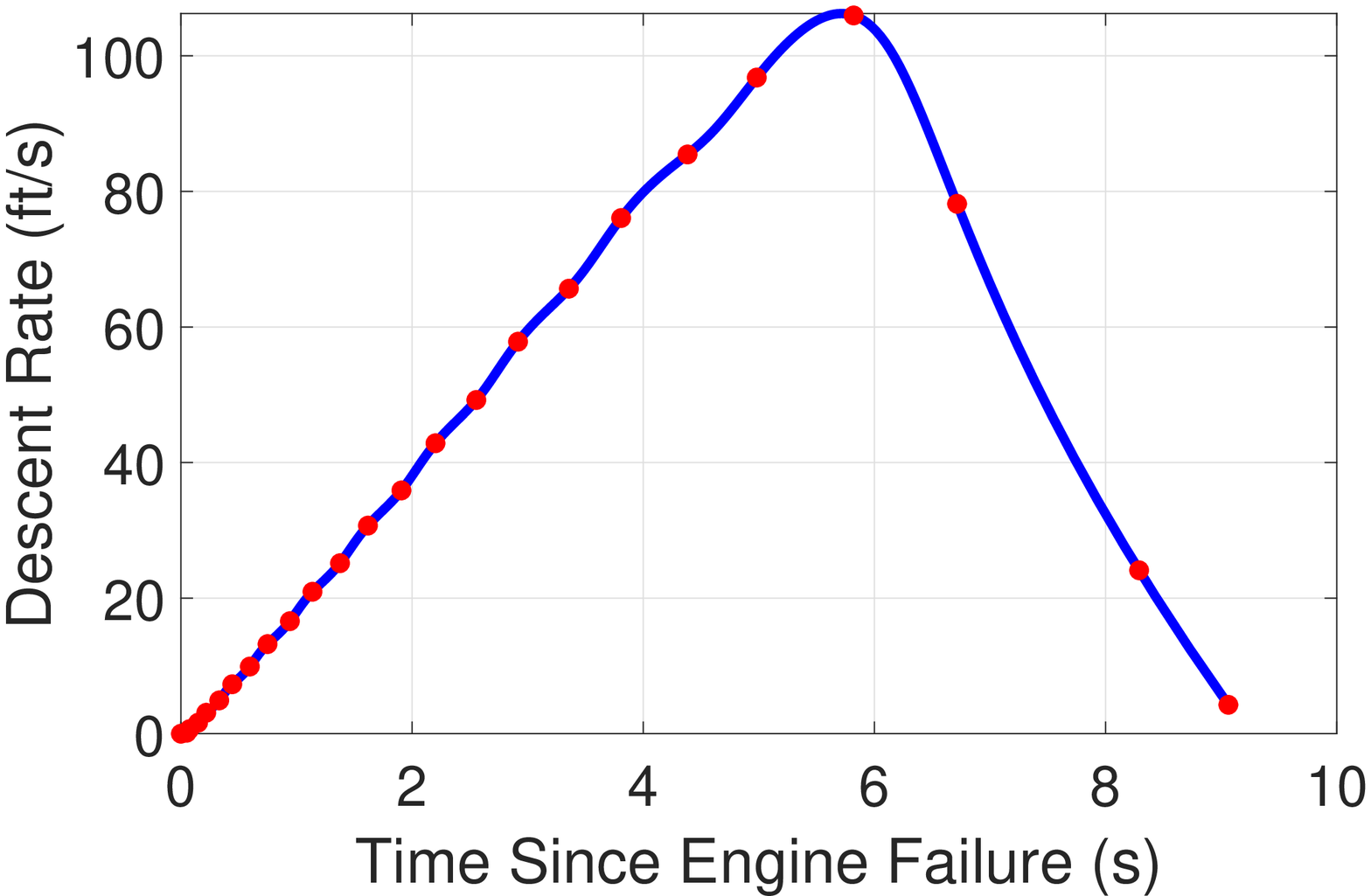}
\par\end{centering}
}\hspace*{\fill}\subfloat[\label{fig:Omega}Rotational speed of the helicopter rotor blades
during landing.]{\centering{}\includegraphics[width=5.5cm]{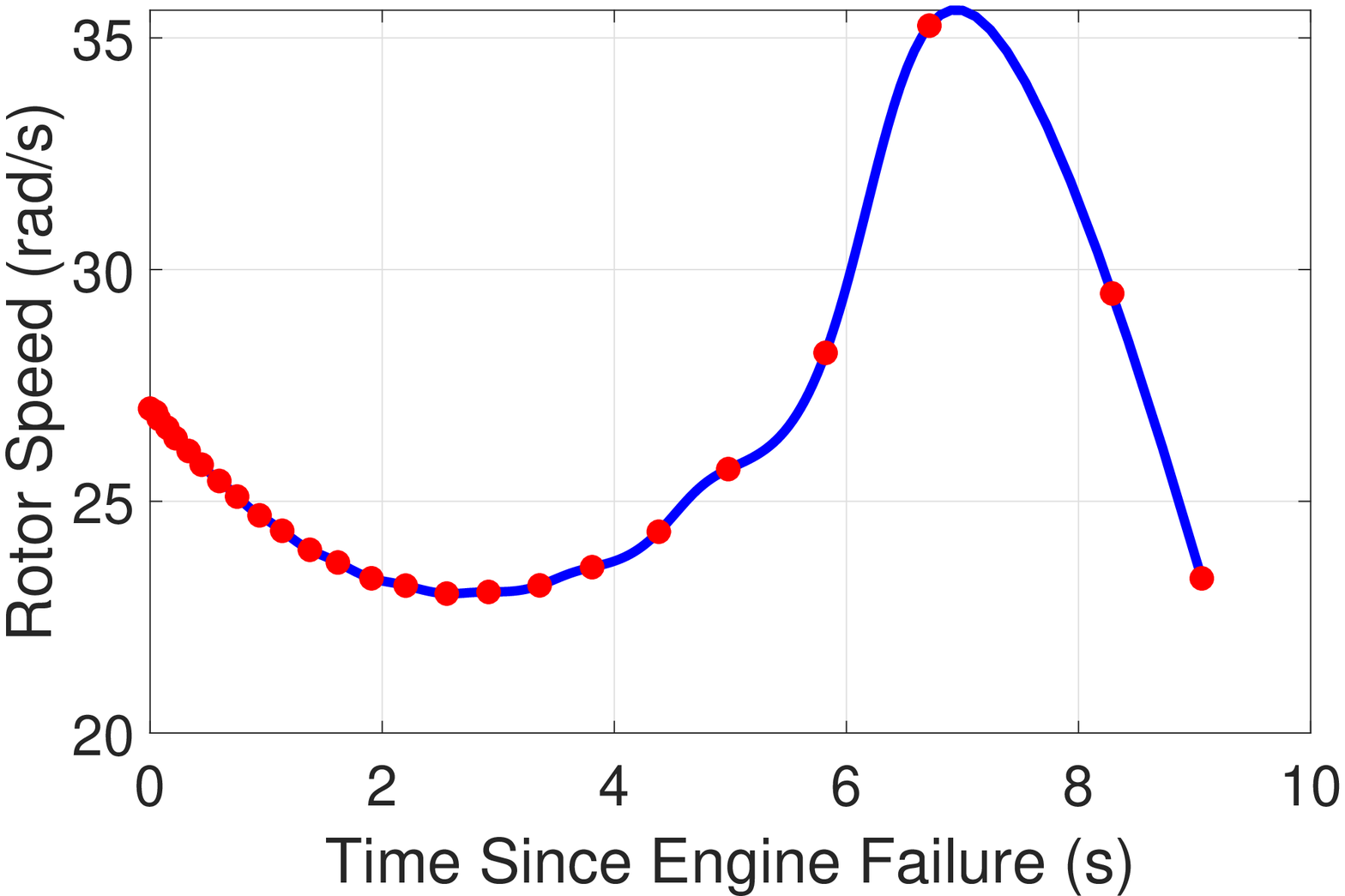}}
\par\end{centering}
\caption{\label{fig:Helicopter-in-autorotation}Selected state trajectories
of an autorotation after engine failure. Red dots indicate the
points of $X$ as found by the NLP optimization. The blue line represents
the Lagrange polynomial approximation, $\zeta^{N}$, of the time-transformed
state trajectory.}
\end{figure*}

The arguments that minimize $\left(\ref{eq:reach tube nlp}\right)$,
denoted as $X^{*}$ and $U^{*}$, are the optimal landing trajectory
and control sequence, respectively, provided that the helicopter is
operating in the reachable tube. As an example we consider a helicopter
in a trimmed cruise at $h=500\,ft$ with a forward speed of $75\,knots$.
Figure \ref{fig:Helicopter-in-autorotation} shows the state evolution
as computed at the time of engine failure. The computed trajectory
results in a safe landing in $9.06\,sec$ as found from $\left(\ref{eq:opt time}\right)$. We note the trajectory of the rotor speed, $\Omega$,
in Figure \ref{fig:Omega}, where an increase in rotor speed is observed
due to translational kinetic energy being transfered into rotational kinetic energy to successfully decelerate,
or ``flare'', at the terminal phase of the landing procedure.

\section{Conclusion and Future Work}

Presented is a numeric method for computing the backwards reachable
tube for general non-linear systems. We are motivated by locating
the unsafe operating states of a helicopter so they can be avoided
by the pilot. By remaining outside of the unsafe set, it ensures it
is possible to initiate an autorotation sequence after engine failure
and guide the helicopter to a safe landing. The method can additionally
be used to compute control inputs in order to complete an autonomous
landing via autorotation. Future work includes augmenting $\left(\ref{eq: heli eom}\right)$
with human pilot dynamics as in \cite{mcruer1974mathematical} and
include pilot reaction delay (\cite{kirchner2019timedelay}). Additionally,
we will investigate integrating failure detection to initiate the
autonomous autorotation landing procedure. 

\balance

\bibliography{root}             

\begin{thebibliography}{31}
\providecommand{\natexlab}[1]{#1}
\providecommand{\url}[1]{\texttt{#1}}
\providecommand{\urlprefix}{URL }
\expandafter\ifx\csname urlstyle\endcsname\relax
  \providecommand{\doi}[1]{doi:\discretionary{}{}{}#1}\else
  \providecommand{\doi}{doi:\discretionary{}{}{}\begingroup
  \urlstyle{rm}\Url}\fi

\bibitem[{FAA(2019)}]{FAAHandbook}
 (2019).
\newblock \emph{Rotorcraft Flying Handbook, {FAA} {Manual} {H-8083-21B}}.
\newblock Federal Aviation Administration.

\bibitem[{Andersson et~al.(2019)Andersson, Gillis, Horn, Rawlings, and
  Diehl}]{Andersson2019}
Andersson, J.A.E., Gillis, J., Horn, G., Rawlings, J.B., and Diehl, M. (2019).
\newblock {CasADi} -- {A} software framework for nonlinear optimization and
  optimal control.
\newblock \emph{Mathematical Programming Computation}, 11(1), 1--36.
\newblock \doi{10.1007/s12532-018-0139-4}.

\bibitem[{Bansal et~al.(2017)Bansal, Chen, Herbert, and
  Tomlin}]{bansal2017hamilton}
Bansal, S., Chen, M., Herbert, S., and Tomlin, C.J. (2017).
\newblock {Hamilton-Jacobi} reachability: {A} brief overview and recent
  advances.
\newblock In \emph{IEEE 56th Annual Conference on Decision and Control (CDC)},
  2242--2253. IEEE.

\bibitem[{Bibik and Narkiewicz(2012)}]{bibik2012helicopter}
Bibik, P. and Narkiewicz, J. (2012).
\newblock Helicopter optimal control after power failure using comprehensive
  dynamic model.
\newblock \emph{Journal of Guidance, Control, and Dynamics}, 35(4), 1354--1362.

\bibitem[{Carlson et~al.(2006)Carlson, Xue, Keane, and Kevin}]{carlson2006h}
Carlson, E.B., Xue, S., Keane, J., and Kevin, B. (2006).
\newblock H-1 upgrades height-velocity diagram development through flight test
  and trajectory optimization.
\newblock In \emph{Annual Forum Proceedings-American Helicopter Society},
  volume~62, 729. American Helicopter Society.

\bibitem[{Chen and Hindson(1986)}]{chen1986influence}
Chen, R.T. and Hindson, W.S. (1986).
\newblock Influence of dynamic inflow on the helicopter vertical response.
\newblock Technical Report 88327, National Aeronautics and Space
  Administration.

\bibitem[{Darbon and Osher(2016)}]{darbon2016algorithms}
Darbon, J. and Osher, S. (2016).
\newblock Algorithms for overcoming the curse of dimensionality for certain
  {Hamilton}-{Jacobi} equations arising in control theory and elsewhere.
\newblock \emph{Research in the Mathematical Sciences}, 3(1), 19.

\bibitem[{Elnagar et~al.(1995)Elnagar, Kazemi, and
  Razzaghi}]{elnagar1995pseudospectral}
Elnagar, G., Kazemi, M.A., and Razzaghi, M. (1995).
\newblock The pseudospectral {Legendre} method for discretizing optimal control
  problems.
\newblock \emph{IEEE Transactions on Automatic Control}, 40(10), 1793--1796.

\bibitem[{Evans(2010)}]{evans10}
Evans, L.C. (2010).
\newblock \emph{Partial Differential Equations}.
\newblock American Mathematical Society, Providence, R.I.

\bibitem[{Garg et~al.(2011)Garg, Hager, and Rao}]{garg2011pseudospectral}
Garg, D., Hager, W.W., and Rao, A.V. (2011).
\newblock Pseudospectral methods for solving infinite-horizon optimal control
  problems.
\newblock \emph{Automatica}, 47(4), 829--837.

\bibitem[{Garg et~al.(2010)Garg, Patterson, Hager, Rao, Benson, and
  Huntington}]{garg2010unified}
Garg, D., Patterson, M., Hager, W.W., Rao, A.V., Benson, D.A., and Huntington,
  G.T. (2010).
\newblock A unified framework for the numerical solution of optimal control
  problems using pseudospectral methods.
\newblock \emph{Automatica}, 46(11), 1843--1851.

\bibitem[{Harno and Kim(2018)}]{harno2018safe}
Harno, H.G. and Kim, Y. (2018).
\newblock Safe flight envelope estimation for rotorcraft: A reachability
  approach.
\newblock In \emph{18th International Conference on Control, Automation and
  Systems (ICCAS)}, 998--1002. IEEE.

\bibitem[{Hopf(1965)}]{hopf1965generalized}
Hopf, E. (1965).
\newblock Generalized solutions of non-linear equations of first order.
\newblock \emph{Journal of Mathematics and Mechanics}, 14, 951--973.

\bibitem[{Johnson(1977)}]{johnson1977helicopter}
Johnson, W. (1977).
\newblock Helicopter optimal descent and landing after power loss.
\newblock Technical Report NASA-TM-X-73244, National Aeronautics and Space
  Administration.

\bibitem[{Kirchner(2019)}]{kirchner2019timedelay}
Kirchner, M.R. (2019).
\newblock A level set approach to online sensing and trajectory optimization
  with time delays.
\newblock In \emph{IFAC-PapersOnline}, volume~52, 301--306.

\bibitem[{Kirchner et~al.(2020)Kirchner, Debord, and Hespanha}]{kirchner2020}
Kirchner, M.R., Debord, M., and Hespanha, J.P. (2020).
\newblock A {Hamilton}-{Jacobi} formulation for optimal coordination of
  heterogeneous multiple vehicle systems.
\newblock \emph{arXiv preprint arXiv:2003.05792}.

\bibitem[{Kirchner et~al.(2018{\natexlab{a}})Kirchner, Hewer, Darbon, and
  Osher}]{kirchner2018primal}
Kirchner, M.R., Hewer, G., Darbon, J., and Osher, S. (2018{\natexlab{a}}).
\newblock A primal-dual method for optimal control and trajectory generation in
  high-dimensional systems.
\newblock In \emph{2018 IEEE Conference on Control Technology and Applications
  (CCTA)}, 1583--1590. IEEE.

\bibitem[{Kirchner et~al.(2018{\natexlab{b}})Kirchner, Mar, Hewer, Darbon,
  Osher, and Chow}]{kirchner2017time}
Kirchner, M.R., Mar, R., Hewer, G., Darbon, J., Osher, S., and Chow, Y.T.
  (2018{\natexlab{b}}).
\newblock Time-optimal collaborative guidance using the generalized {Hopf}
  formula.
\newblock \emph{IEEE Control Systems Letters}, 2(2), 201--206.

\bibitem[{Lee(1985)}]{lee1986optimal}
Lee, A.Y.N. (1985).
\newblock \emph{Optimal Landing of a Helicopter in Autorotation}.
\newblock Ph.D. thesis, Department of Aeronautics and Astronautics. Stanford
  University.

\bibitem[{Lee et~al.(1988)Lee, Bryson, and Hindson}]{lee1988optimal}
Lee, A.Y., Bryson, A.E., and Hindson, W.S. (1988).
\newblock Optimal landing of a helicopter in autorotation.
\newblock \emph{Journal of Guidance, Control, and Dynamics}, 11(1), 7--12.

\bibitem[{McRuer and Krendel(1974)}]{mcruer1974mathematical}
McRuer, D.T. and Krendel, E.S. (1974).
\newblock Mathematical models of human pilot behavior.
\newblock AGARDDograph 188, Advisory Group on Aerospace Research and
  Development.

\bibitem[{Mitchell et~al.(2005)Mitchell, Bayen, and Tomlin}]{mitchell2005time}
Mitchell, I., Bayen, A.M., and Tomlin, C.J. (2005).
\newblock A time-dependent {Hamilton}-{Jacobi} formulation of reachable sets
  for continuous dynamic games.
\newblock \emph{IEEE Transactions on Automatic Control}, 50(7), 947--957.

\bibitem[{Mitchell(2007{\natexlab{a}})}]{Mitchell_toolbox}
Mitchell, I.M. (2007{\natexlab{a}}).
\newblock A toolbox of level set methods.
\newblock Technical Report TR-2007-11, UBC Department of Computer Science.

\bibitem[{Mitchell(2008)}]{mitchell2008flexible}
Mitchell, I.M. (2008).
\newblock The flexible, extensible and efficient toolbox of level set methods.
\newblock \emph{Journal of Scientific Computing}, 35(2), 300--329.

\bibitem[{Mitchell(2007{\natexlab{b}})}]{mitchell2007comparing}
Mitchell, I.M. (2007{\natexlab{b}}).
\newblock Comparing forward and backward reachability as tools for safety
  analysis.
\newblock In \emph{International Workshop on Hybrid Systems: Computation and
  Control}, 428--443. Springer.

\bibitem[{Osher and Fedkiw(2006)}]{osher2006level}
Osher, S. and Fedkiw, R. (2006).
\newblock \emph{Level Set Methods and Dynamic Implicit Surfaces}, volume 153.
\newblock Springer Science \& Business Media.

\bibitem[{Ross and Karpenko(2012)}]{ross2012review}
Ross, I.M. and Karpenko, M. (2012).
\newblock A review of pseudospectral optimal control: From theory to flight.
\newblock \emph{Annual Reviews in Control}, 36(2), 182--197.

\bibitem[{Shen et~al.(2011)Shen, Tang, and Wang}]{shen2011spectral}
Shen, J., Tang, T., and Wang, L.L. (2011).
\newblock \emph{Spectral Methods: Algorithms, Analysis and Applications},
  volume~41.
\newblock Springer Science \& Business Media.

\bibitem[{Subbotin(1995)}]{subbotin1995generalized}
Subbotin, A.I. (1995).
\newblock \emph{Generalized Solutions of First Order {PDE}s: The Dynamical
  Optimization Perspective}.
\newblock Birkh\"{a}user.

\bibitem[{W{\"a}chter and Biegler(2006)}]{wachter2006implementation}
W{\"a}chter, A. and Biegler, L.T. (2006).
\newblock On the implementation of an interior-point filter line-search
  algorithm for large-scale nonlinear programming.
\newblock \emph{Mathematical Programming}, 106(1), 25--57.

\bibitem[{Yomchinda(2013)}]{yomchinda2013real}
Yomchinda, T. (2013).
\newblock \emph{Real-time Path Planning and Autonomous Control for Helicopter
  Autorotation}.
\newblock Ph.D. thesis, Department of Aerospace Engineering, Pennsylvania State
  University.

\end{thebibliography}

                                                   







\end{document}